\documentclass[10pt,twocolumn,letterpaper]{article}

\usepackage{cvpr}              % To produce the CAMERA-READY version
\usepackage{graphicx}
\usepackage{amsmath}
\usepackage{amssymb}
\usepackage{booktabs}
\usepackage{multirow}
\usepackage{bm}
\usepackage{pifont}% http://ctan.org/pkg/pifont
\usepackage[pagebackref,breaklinks,colorlinks]{hyperref}

\newcommand{\cmark}{\ding{51}}%
\newcommand{\xmark}{\ding{55}}%

\usepackage[capitalize]{cleveref}
\crefname{section}{Sec.}{Secs.}
\Crefname{section}{Section}{Sections}
\Crefname{table}{Table}{Tables}
\crefname{table}{Tab.}{Tabs.}

\def\cvprPaperID{2082} % *** Enter the CVPR Paper ID here
\def\confName{CVPR}
\def\confYear{2022}

\begin{document}
	
	\title{Video Frame Interpolation with Transformer}
	
	\author{
		Liying Lu\textsuperscript{1} \quad Ruizheng Wu\textsuperscript{2} \quad Huaijia Lin\textsuperscript{2} \quad  Jiangbo Lu\textsuperscript{2} \quad Jiaya Jia\textsuperscript{1}\\
		$^1$ The Chinese University of Hong Kong \qquad
		$^2$ SmartMore\\
		{\tt\small \{lylu,rzwu,linhj,leojia\}@cse.cuhk.edu.hk, jiangbo@smartmore.com} 
	}

	\maketitle
	
	%%%%%%%%% ABSTRACT
	\begin{abstract}
		Video frame interpolation~(VFI), which aims to synthesize intermediate frames of a video, has made remarkable progress with development of deep convolutional networks over past years. Existing methods built upon convolutional networks generally face challenges of handling large motion due to the locality of convolution operations. To overcome this limitation, we introduce a novel framework, which takes advantage of Transformer to model long-range pixel correlation among video frames. Further, our network is equipped with a novel cross-scale window-based attention mechanism, where cross-scale windows interact with each other. This design effectively enlarges the receptive field and aggregates multi-scale information. Extensive quantitative and qualitative experiments demonstrate that our method achieves new state-of-the-art results on various benchmarks. The source code is available at \href{https://github.com/dvlab-research/VFIformer}{https://github.com/dvlab-research/VFIformer}.
		%the effectiveness of our proposed model.
		%and endows our framework with a strong capability of modeling long-range dependencies. 
	\end{abstract}
	
	%%%%%%%%% BODY TEXT

	\section{Introduction}
	\label{sec:intro}

	Video frame interpolation~(VFI) is a fundamental video processing task in which intermediate frames are synthesized between given consecutive ones to increase the frame rate. It is effective in alleviating motion blur and judder, and has become a compelling strategy for numerous applications, such as novel view synthesis~\cite{flynn2016deepstereo,kalantari2016learning}, video compression~\cite{wu2018video}, video restoration~\cite{haris2020space,xiang2020zooming,kim2020fisr}, and slow motion generation~\cite{niklaus2017video,liu2017video,jiang2018super,niklaus2018context,peleg2019net,bao2019depth}.
	Many popular algorithms adopt optical flow warping~\cite{liu2017video,jiang2018super,bao2019depth,niklaus2018context,softsplat,bao2019memc,liu2019deep,xu2019quadratic,park2020bmbc,asymmetric,xvfi} to tackle this challenging task. Though achieving remarkable performance, these methods built upon convolutional networks generally face challenges of capturing long-range spatial interactions due to the intrinsic locality of convolution operations, thus limited in handling large motion, which is one of the main challenges of VFI. 
	%Besides, the quality of their generated results generally depends on the accuracy of the estimated flows to a large extent, especially when facing situations with occlusions, blur, and abrupt brightness changes.
	
\begin{figure}[t]
		\begin{center}
			\includegraphics[width=1.0\linewidth]{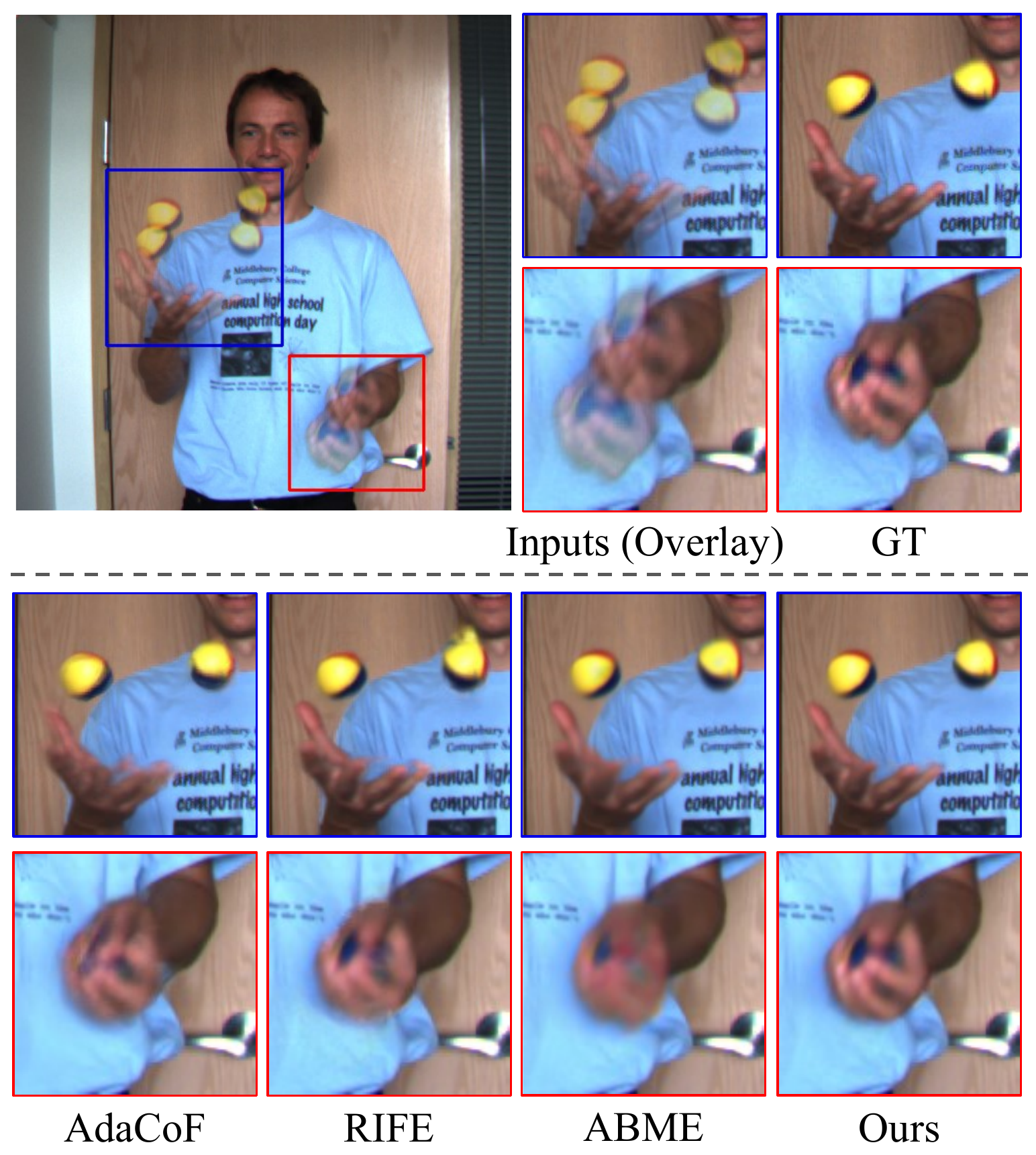}
		\end{center}
		%\vspace{-0.18in}
		\caption{Visual comparison on a challenging sample from Middlebury dataset~\cite{middlebury}. Our method produces the result more appealing than the three leading VFI methods, \ie, AdaCoF~\cite{lee2020adacof}, RIFE~\cite{rife} and ABME~\cite{asymmetric}.}
		\label{fig:teaser}
	\end{figure}

	Recently, natural language processing~(NLP)~\cite{vaswani2017attention,devlin2018bert,brown2020language} and computer vision~\cite{dosovitskiy2020image,carion2020end,liu2021swin} tasks achieve notable progress using Transformers, which is a highly adaptive architecture with strong modeling capability. In this work, we are inspired to explore the application of Transformers in the context of video frame interpolation and introduce a novel network, VFIformer. With the attention mechanism as the core operation, VFIformer is able to model pixel correspondence between different frames. Besides, its strong capability of capturing long-range dependency is helpful for handling large motion~(see Fig.~\ref{fig:teaser}).

	Since the vanilla Transformer needs high memory and computational cost, the proposed VFIformer is designed in a UNet~\cite{unet} architecture where features are processed in different scales to reduce the computational complexity and enlarge the receptive field. Besides, to overcome the quadratic complexity, inspired by recent work~\cite{liu2021swin,wang2021uformer,liang2021swinir,chu2021twins}, our VFIformer is built upon window-based attention where feature maps are divided into non-overlapping sub-windows. Self-attention is only performed within each sub-window. Despite computationally efficient, such an approach prohibits information interaction between different windows and leads to limited receptive field. We address this problem by proposing cross-scale window-based attention, where the attention is computed among feature windows of different scales. 
	
	Our design enjoys two merits. (1) Compared with windows at the original scale, the corresponding windows at coarser scales cover more content. As a result, the interaction among these windows effectively enlarges the receptive field. (2) Features at coarser scales naturally contain smaller displacement and thus provide informative motion priors for the original scale to facilitate synthesis.

	Our contributions are summarized as follows:
	\begin{itemize}
		\item We propose a novel framework integrated with the Transformer for the VFI task, which takes advantage of the Transformer to model long-range pixel correlations among the video frames.
		\item A cross-scale window-based attention mechanism is introduced to enlarge the receptive field of current window-based attention to adapt to the challenges of large motions in the VFI task.
		\item Our model achieves state-of-the-art performance for the VFI task on multiple public benchmarks.
	\end{itemize}

	%We build upon the recent success of transformer xxx
	%Recent developments in transformer may provide an opportunity for a new approach in depth estimation.
	
	%-------------------------------------------------------------------------
	\section{Related Work}
	
	\subsection{Video Frame Interpolation}
	
	Video frame interpolation, aiming to synthesize intermediate frames between existing ones of a video, is a long-standing problem. Existing methods can be classified into three categories. They are phase-based, kernel-based, and motion-based ones. 
	
	\textit{Phase-based} methods represent motion in the phase shift of individual pixels. They interpolate phase and amplitude across the levels of a multi-scale pyramid through optimization~\cite{meyer2015phase} or neural networks~\cite{meyer2018phasenet}. A common drawback of these approaches is that they are only applicable to limited-range motion.
	
	\textit{Kernel-based} methods jointly perform motion estimation and motion compensation in a single step. Niklaus~\etal~\cite{niklaus2017videoaa} estimate a spatially-adaptive convolution kernel for each pixel using a convolutional network. The intermediate frame is then generated by convolving the input frames with the predicted kernels. Further development in this field includes using adaptive separable convolutions~\cite{niklaus2017video} to reduce network parameters, adopting deformable convolution or its alternatives to estimate both kernel weights and offset vectors~\cite{lee2020adacof,shi2020video,cheng2021multiple}, integrating optical flow and interpolation kernels together~\cite{bao2019memc,bao2019depth} to improve the performance, and developing loss functions that combine adaptive convolution and trilinear interpolation~\cite{peleg2019net}. 
	
	These methods tend to yield blurry results when handling fast-moving objects since they hallucinate pixel values directly. Besides, to handle large motion, the estimated kernels are designed to be large, leading to a large number of parameters to learn.
	
	For \textit{motion-based} methods, optical flow is estimated to warp the input frames. Liu~\etal~\cite{liu2017video} introduce a deep network that produces 3D optical flow vectors across space and time, and warps input frames by trilinear sampling. Jiang~\etal~\cite{jiang2018super} linearly combine optical flow between the input images to approximate the intermediate flow. 
	
	Recent work has explored a few strategies for improving the performance of such methods. These efforts include utilizing additional contextual information to interpolate high-quality results~\cite{niklaus2018context}, developing unsupervised techniques by cycle consistency~\cite{reda2019unsupervised}, detecting the occlusion by exploring the depth information~\cite{bao2019depth}, forward-warping input frames using softmax splatting~\cite{softsplat}, using quadratic interpolation to overcome the limitation of linear models~\cite{xu2019quadratic,liu2020enhanced}, leveraging the distillation loss to supervise the intermediate flows~\cite{rife}, and constructing efficient architectures for large resolution images~\cite{sim2021xvfi,choi2021motion}. We note that methods built upon convolutional networks generally face challenges of modeling long-term dependencies thus limiting large motion handling. 
	%Besides, they are also limited in dealing with situations with occlusions, blur, and abrupt brightness changes.
	
	\begin{figure*}[t]
		\begin{center}
			\includegraphics[width=1.0\linewidth]{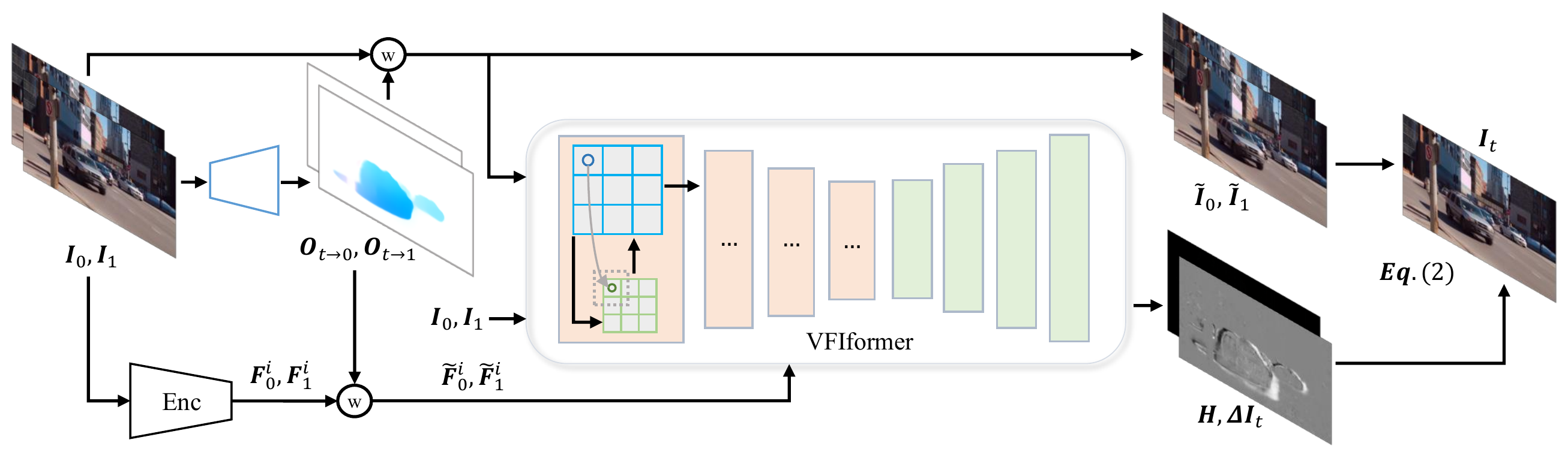}
		\end{center}
		%\vspace{-0.18in}
		\caption{Overview of our proposed framework. At first, a convolutional network is used to directly estimate the intermediate optical flows $\bm{O}_{t\rightarrow0}$ and $\bm{O}_{t\rightarrow1}$. An encoder $Enc$ is used to extract multi-scale features $\bm{F}_0^i$ and $\bm{F}_i^i$ from the input frames, where $i=0,1,2,3$. The input frames and extracted features are then backward warped by the estimated flow, producing $\widetilde{\bm{I}}_{0}$, $\widetilde{\bm{I}}_{1}$, $\widetilde{\bm{F}}_{0}^{i}$, and $\widetilde{\bm{F}}_{1}^{i}$. At last, to generate the final results, the input frames and the warped features are fed into the proposed VFIformer, in which cross-scale attention is employed to enlarge the receptive field.}
		\label{fig:overview}
	\end{figure*}

	\subsection{Transformer}
	
	Transformer was first proposed by Vaswani~\etal~\cite{vaswani2017attention} for machine translation. It consists of stacked self-attention layers for modeling dense relation among input tokens and has shown great flexibility. After breakthrough with the advent of Transformer in NLP, research of Transformer in computer vision becomes popular. Carion~\etal~\cite{carion2020end} propose an end-to-end detection Transformer~(DETR) for direct set prediction. Dosovitskiy~\etal~\cite{dosovitskiy2020image} propose ViT, which is a pure Transformer for image classification and achieves decent results. Liu~\etal~\cite{liu2021swin} present a general-purpose backbone, called Swin Transformer, which achieves linear computational complexity by computing self-attention within non-overlapping windows. A shifted window scheme is also proposed for cross-window connection. 
	
	Apart from high-level vision tasks, several attempts have also been made to integrate the Transformer into low-level vision tasks~\cite{ttsr,cao2021video,chen2021pre,liang2021swinir}. Chen~\etal~\cite{chen2021pre} develop a pre-trained model for image processing using the Transformer architecture. Liang~\etal~\cite{liang2021swinir} propose SwinIR for image restoration based on the Swin Transformer. Cao~\etal~\cite{cao2021video} adapt Transformer for video super-resolution, and an optical flow-based feed-forward layer is integrated for feature alignment. In this work, we introduce the Transformer into the VFI task, aiming to leverage its power of capturing long-range correlation.

	%------------------------------------------------------------------------

	\section{Our Method}

	Given two input frames $\bm{I}_0$ and $\bm{I}_1$, video frame interpolation is to synthesize an intermediate frame $\bm{I}_t$. Our framework is illustrated in Fig.~\ref{fig:overview}. At first, we utilize a convolutional network~(called flow estimator in the following) and an encoder $Enc$ to obtain the preliminary elements, including optical flows $\bm{O}_{t\rightarrow0}$ and $\bm{O}_{t\rightarrow1}$, corresponding warped features $\widetilde{\bm{F}}_{0}^{i}$ and $\widetilde{\bm{F}}_{1}^{i}$, and warped images $\widetilde{\bm{I}}_{0}$ and $\widetilde{\bm{I}}_{1}$. 
	
	With these preliminary results as input, VFIformer (Sec.~\ref{sec:vfiformer}) is utilized to capture long-range pixel interaction, generating the mask and residual for final synthesis. To enlarge the receptive field of the window-based attention in VFIformer, we design a cross-scale window-based attention (Sec.~\ref{sec:csma}) mechanism to make trade-off between efficiency and performance.

	\subsection{VFIformer}
	\label{sec:vfiformer}
	
	\begin{figure*}[t]
		\begin{center}
			\includegraphics[width=0.8\linewidth]{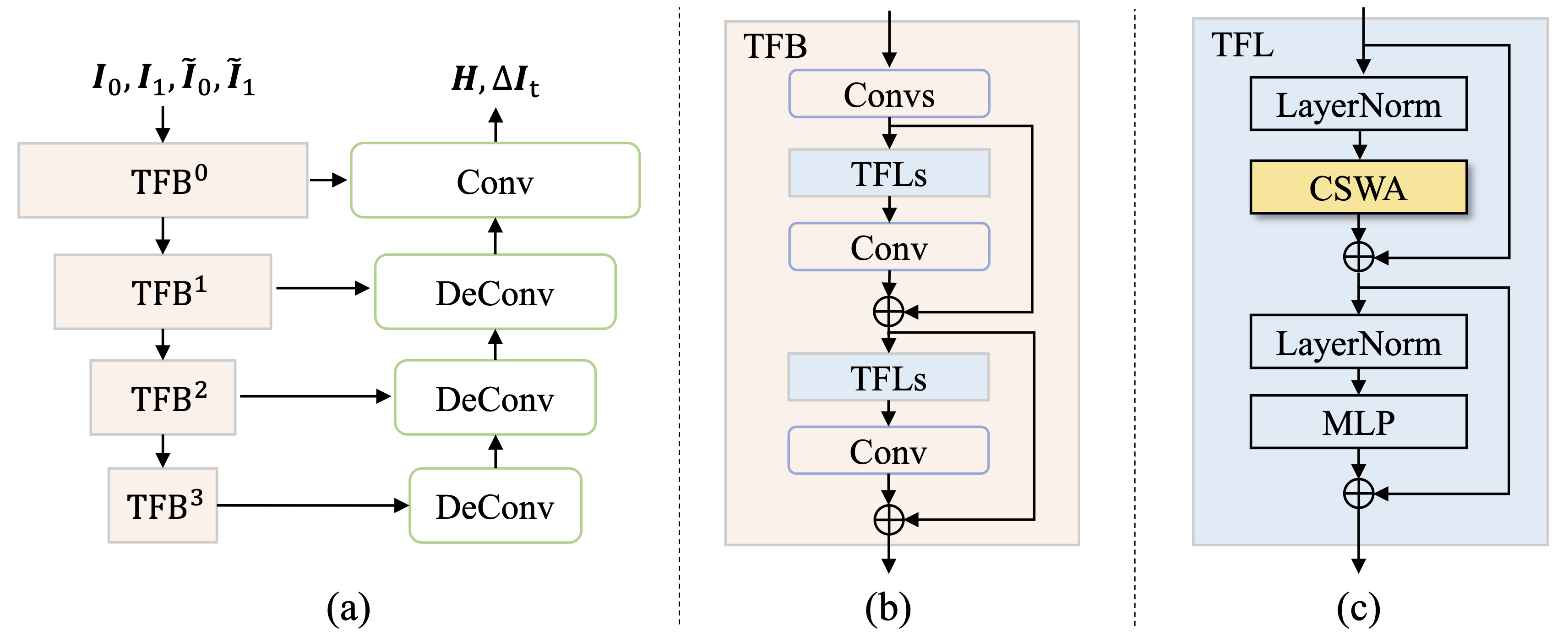}
		\end{center}
		%\vspace{-0.18in}
		\caption{Structure of the proposed VFIformer. (a) VFIformer is designed in a UNet architecture. Its encoder consists of several Transformer blocks~(TFBs). (b) Structure of the Transformer blocks~(TFB). (c) Structure of the Transformer layers~(TFL), where Cross-Scale Window-based Attention (CSWA) is a key component.}
		\label{fig:fe_transformer}
		\vspace{-0.1in}
	\end{figure*}

	Since the locality of convolution constrains its receptive field, previous VFI methods built upon convolutional networks generally face challenges of capturing long-range spatial interactions. In contrast, our work builds upon the recent advance that integrates Transformers into vision models to learn long-range dependencies. We propose the VFIformer, which is able to aggregate information over large receptive fields and is effective in handling large  displacement.
	
	As shown in Fig.~\ref{fig:fe_transformer}a, our VFIformer is designed in a UNet architecture, where features are processed in different scales to reduce computational complexity and enlarge the receptive field.
	The encoder of VFIformer consists of several Transformer blocks~(TFB), and the decoder consists of standard convolutions and transposed convolutions.
	For the $i$-th TFB, its output feature $F_t^i$ is produced as
	\begin{align}
	\bm{F}_t^i = {TFB}^{i}([\bm{F}_t^{i-1}, \widetilde{\bm{F}}_{0}^{i}, \widetilde{\bm{F}}_{1}^{i}]) \;,
	\end{align}
	where $\bm{F}_t^{i-1}$ is the feature from the last TFB, $\widetilde{\bm{F}}_{0}^{i}$ and $\widetilde{\bm{F}}_{1}^{i}$ are the features from $Enc$ warped by the intermediate optical flow $\bm{O}_{t\rightarrow0}$ and $\bm{O}_{t\rightarrow1}$. ${TFB}^{i}$ denotes the operations of the $i$-th TFB. The first TFB takes as input the concatenation of frames $\bm{I}_0$, $\bm{I}_1$, $\widetilde{\bm{I}}_{0}$, and $\widetilde{\bm{I}}_{1}$ without features. 
	
	At last, the decoder of VFIformer produces a soft mask $\bm{H}$ and an image residual $\Delta \bm{I}_t$ to synthesize the final intermediate frame $\bm{I}_{t}$ as
	\begin{align}
	\bm{I}_{t} = \bm{H} \odot \widetilde{\bm{I}}_{0} + (1-\bm{H}) \odot \widetilde{\bm{I}}_{1} + \Delta{\bm{I}_t} \;,
	\end{align} 
	where $\odot$ denotes the Hadamard product. The soft mask is used to blend the two warped frames $\widetilde{\bm{I}}_{0}$, $\widetilde{\bm{I}}_{1}$. The residual is used to compensate flow errors and occlusion.

	We then look into the detailed structures of TFB. As illustrated in Fig.~\ref{fig:fe_transformer}(b), each TFB consists of several Transformer layers TFL (see Fig.~\ref{fig:fe_transformer}(c)) and convolutional layers. 
	The features in the $l$-th TFL are processed as
	\begin{align}
	\widehat{\bm{z}}^{l} = CSWA(LN(\bm{z}^{l-1})) + \bm{z}^{l-1} \;, \\
	\bm{z}^{l} = MLP(LN(\widehat{\bm{z}}^{l})) + \widehat{\bm{z}}^{l},
	\end{align}
	where $\bm{z}^{l-1}$ is the feature generated by the $(l-1)$-th TFL. LN and MLP denote the LayerNorm and Multi-Layer Perceptrons. CSWA denotes our proposed cross-scale window-based attention, which is explained in the following.

	\subsection{Cross-Scale Window-based Attention~(CSWA)}
	\label{sec:csma}
	
	\paragraph{Window-based Attention~(WA)}
	We first revisit the window-based attention method.
	Though the key component self-attention of Transformer, has shown great flexibility and strong modeling capability, a known fact is that its power comes at the price of high computational complexity. Inspired by \cite{liu2021swin,liang2021swinir}, we employ window-based attention~(WA) to reduce the computational cost, where feature maps are divided into sub-windows. Self-attention is only performed within each sub-window. Specifically, for a feature map $\bm{F} \in \mathbb{R}^{H\times W\times C}$, we divide it into $\frac{HW}{M^2}$ sub-windows of size $M\times M$.
	Taking one of the windows $\bm{X}\in \mathbb{R}^{(M^2, C)}$ as an example, its \textit{query}, \textit{key} and \textit{value} matrices $\bm{Q}$, $\bm{K}$ and $\bm{V}\in \mathbb{R}^{(M^2, d)}$ are computed as
	\begin{align}
	\bm{Q}=\bm{X}\bm{W}_Q,~~ \bm{K}=\bm{X}\bm{W}_{K},~~ \bm{V}=\bm{X}\bm{W}_{V} \;, 
	\label{eq:qkv}
	\end{align}
	where $\bm{W}_Q$, $\bm{W}_{K}$, and $\bm{W}_{V}$ are projection matrices shared across different windows. Afterwards, the self-attention is computed as
	\begin{align}
	Attn(\bm{Q},\bm{K},\bm{V})=Softmax(\frac{\bm{Q}\bm{K}^{T}}{\sqrt{d}}+\bm{P})\bm{V} \;,
	\label{eq:attn}
	\end{align}
	where $\bm{P}$ is the learnable positional encoding, and $d$ is the $query/key$ dimension.
	
	\begin{figure}[t]
		\begin{center}
			\includegraphics[width=1.0\linewidth]{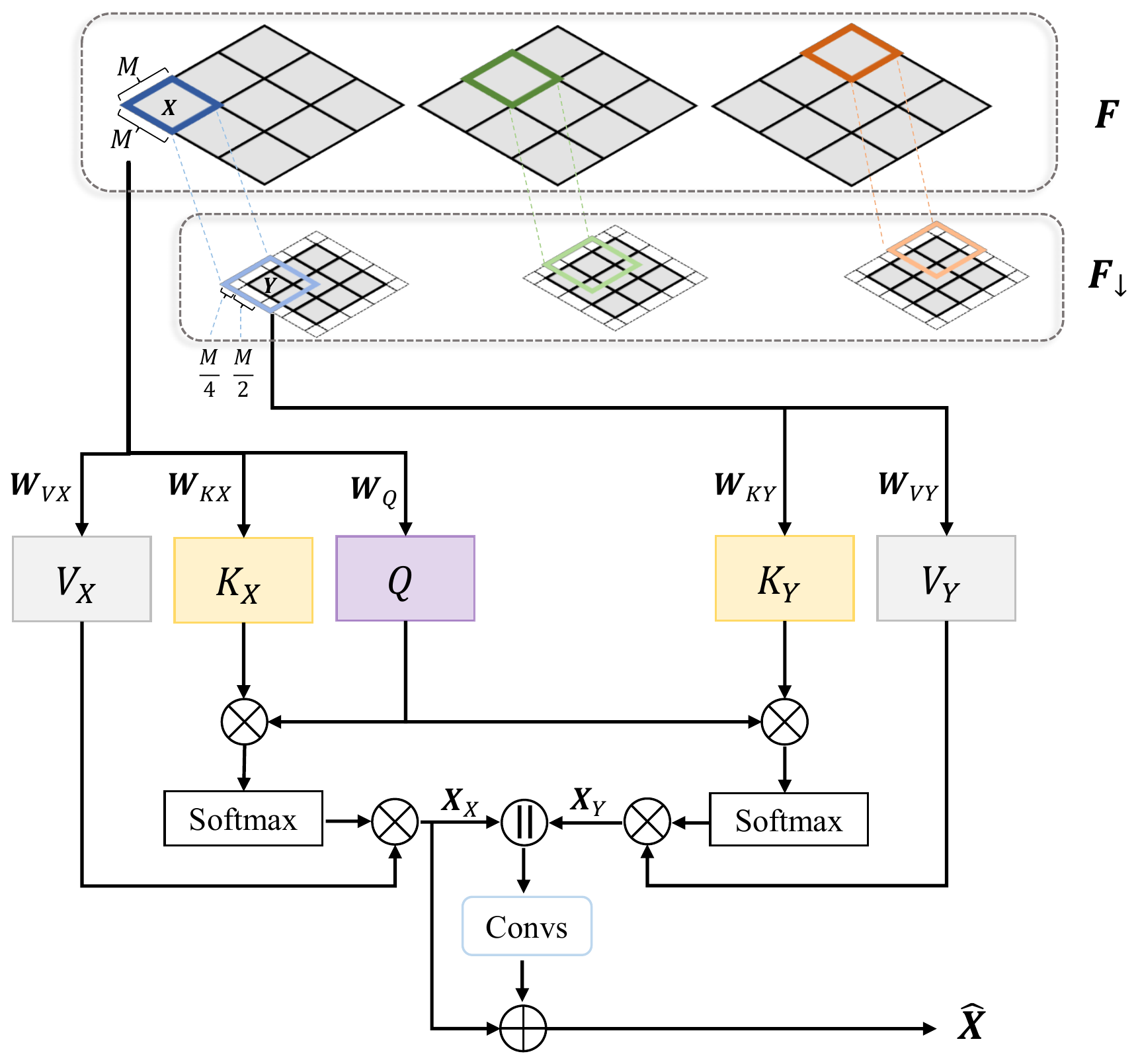}
		\end{center}
		\vspace{-0.1in}
		\caption{Proposed cross-scale window-based attention~(CSWA). The fine-scale feature $\bm{F}$ is divided into non-overlapping windows~(dark-colored block), and the coarse-scale feature $\bm{F}_{\downarrow}$ is divided into overlapping windows~(light-colored block). The feature windows with the same color~(\eg, $\bm{X}$ and $\bm{Y}$) interact with each other.}
		\label{fig:msattn}
		\vspace{-0.1in}
	\end{figure}
	
	\paragraph{Cross-Scale Window-based Attention~(CSWA)}
	Although window-based attention is computationally efficient, the drawback is that the receptive field is still limited, resulting in limited information interaction between different windows. We address this by introducing the cross-scale window-based attention~(CSWA), which enlarges the receptive field in an effective way. 
	
	The details are shown in Fig.~\ref{fig:msattn}, for the input feature map $\bm{F} \in \mathbb{R}^{H\times W\times C}$, we first down-sample it by scale 2 to get $\bm{F}_{\downarrow}\in \mathbb{R}^{\frac{H}{2}\times \frac{W}{2}\times C}$. Then $\bm{F}$ is divided into $\frac{HW}{M^2}$ \textit{non-overlapping} sub-windows, following the same procedure in WA as mentioned above. As for $\bm{F}_{\downarrow}$, we first pad it with padding size $\frac{M}{4}\times \frac{M}{4}$ in the mode of reflection, and then divide it into \textit{overlapping} sub-windows of size $M\times M$. 
	
	Taking one of the windows $\bm{X}\in \mathbb{R}^{(M^2, C)}$ in $\bm{F}$, we denote $\bm{Y}\in \mathbb{R}^{(M^2, C)}$ as its corresponding window in $\bm{F}_{\downarrow}$. Following Eq.~\eqref{eq:qkv}, we calculate the query $\bm{Q}$ only for the window $\bm{X}$ from the original feature $\bm{F}$. As for the \textit{key} and \textit{value}, we calculate them for both the window $\bm{X}$ and window $\bm{Y}$ to interact features in different scales. The procedure is written as
	\begin{align}
	\bm{Q}=\bm{X}\bm{W}_Q \;, \\
	\bm{K}_X=\bm{X}\bm{W}_{KX},~~ \bm{K}_Y=\bm{Y}\bm{W}_{KY} \;, \\
	\bm{V}_X=\bm{X}\bm{W}_{VX},~~ \bm{V}_Y=\bm{Y}\bm{W}_{VY} \;, 
	\end{align}
	where $\bm{W}_Q$, $\bm{W}_{KX}$, $\bm{W}_{KY}$, $\bm{W}_{VX}$ and $\bm{W}_{VY}$ are projection matrices. Afterwards, the attention is computed within the set of $(\bm{Q}, \bm{K}_X, \bm{V}_X)$ and $(\bm{Q}, \bm{K}_Y, \bm{V}_Y)$ in a similar way as Eq.~\eqref{eq:attn}, producing $\bm{X}_X$ and $\bm{X}_Y$. The final result is generated as
	\begin{align}
	\hat{\bm{X}}= \bm{X}_X + Convs([\bm{X}_X, \bm{X}_Y]) \;,
	\end{align}
	where $[~]$ denotes concatenation in the channel dimension.

	As shown in Fig.~\ref{fig:msattn}, windows with the same color of $\bm{F}$ and $\bm{F}_{\downarrow}$ interact with each other, introducing multi-scale information and therefore generating more representative features. On the other hand, windows of $\bm{F}_{\downarrow}$ cover larger context than those of $\bm{F}$. For example, window $\bm{Y}$ in $\bm{F}_{\downarrow}$ actually covers 4 times as much context as the window $\bm{X}$ in $\bm{F}$.
	
\begin{figure}[t]
		\begin{center}
			\includegraphics[width=0.7\linewidth]{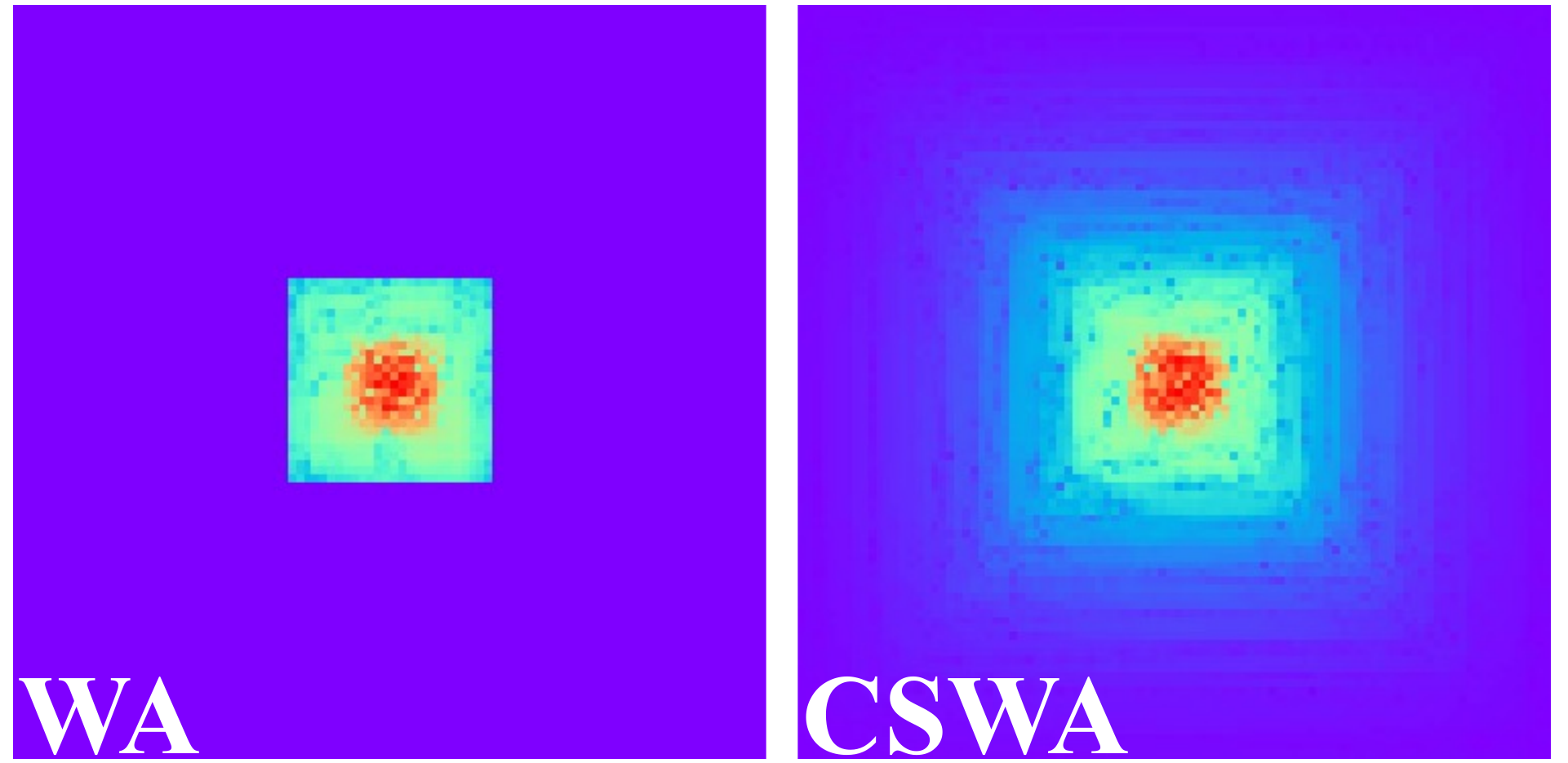}
		\end{center}
		\vspace{-0.1in}
		\caption{The effective receptive field of WA and CSWA.}
		\label{fig:receptive}
		\vspace{-0.1in}
	\end{figure}

	In this way, the receptive field of self-attention is enlarged effectively. We adopt the widely used effective receptive field~(ERF)~\cite{erf} to visualize the ERFs of WA and CSWA. Fig.~\ref{fig:receptive} shows the ERF of a TFB equipped with {\bf left}: WA, {\bf right}: CSWA, it is obvious that the ERF of CSWA is much larger than that of WA.
	% without introducing heavy computational cost. 
	
	%------------------------------------------------------------------------
	
	%------------------------------------------------------------------------
	\subsection{Loss Functions}
	
	\noindent\textbf{Reconstruction loss.} We adopt $L_{1}$ loss as the reconstruction loss as
	\begin{align}
	\mathcal{L}_{rec} = \| {\bm{I}}^{GT}_t - \bm{I}_t \|_{1} \;,
	\end{align}
	where $\bm{I}^{GT}_t$ and $\bm{I}_t$ denote the ground-truth intermediate frame and the generated one.
	
	\vspace{0.05in}
	\noindent\textbf{Census loss.} Census loss~\cite{meister2018unflow,zhong2019unsupervised} $\mathcal{L}_{css}$ is robust to illumination changes, which is defined as the soft Hamming distance between census-transformed~\cite{zabih1994non} image patches of $\bm{I}^{GT}_t$ and $\bm{I}_t$.

	\vspace{0.05in}	
	\noindent\textbf{Distillation loss.} Following \cite{rife}, we use distillation loss to explicitly supervise the estimated flows as
	\begin{align}
	\mathcal{L}_{dis} = \| \bm{O}^{*}_{t\rightarrow0} - \bm{O}_{t\rightarrow0} \|_{1} + \| \bm{O}^{*}_{t\rightarrow1} - \bm{O}_{t\rightarrow1} \|_{1} \;,
	\end{align}
	where $\bm{O}^{*}_{t\rightarrow0}$ and $\bm{O}^{*}_{t\rightarrow1}$ are flows generated by a pretrained flow estimation network~\cite{hui2018liteflownet}. $\bm{O}_{t\rightarrow0}$ and $\bm{O}_{t\rightarrow1}$ are flow estimated by our flow estimator. 
	%Note that we do not apply distillation loss on the flows refined by BLRB.
	
	%RIFE~\cite{rife} has shown that explicitly supervising the estimated flows brings large performance improvement. Thus we also 
	\vspace{0.05in}	
	\noindent\textbf{Full objective.} Our full objective is defined as
	\begin{align}
	\mathcal{L} = {\lambda}_{rec}\mathcal{L}_{rec} + {\lambda}_{css}\mathcal{L}_{css} + {\lambda}_{dis}\mathcal{L}_{dis} \;,
	\end{align}
	where ${\lambda}_{rec}$, ${\lambda}_{css}$ and ${\lambda}_{dis}$ are loss weights for $\mathcal{L}_{rec}$, $\mathcal{L}_{css}$ and $\mathcal{L}_{dis}$, respectively.
	
	%------------------------------------------------------------------------
	\section{Experiments}
	
	\subsection{Datasets}
	
	Our model is trained on the Vimeo90K training set and evaluated on various datasets.

	\vspace{0.05in}	
	\noindent\textbf{Vimeo90K~\cite{vimeo90k}.} The Vimeo90K training set contains $51,312$ triplets, where each triplet consists of three consecutive video frames with resolution $448\times 256$. The Vimeo90K training set contains $3,782$ triplets whose resolution is also $448\times 256$.

	\vspace{0.05in}	
	\noindent\textbf{UCF101~\cite{ucf101}.} It contains videos with a large variety of human actions. There are 379 triplets with a resolution of $256\times 256$.

	\vspace{0.05in}	
	\noindent\textbf{Middlebury.} The Middlebury dataset has two subsets, in which the OTHER dataset provides the ground-truth intermediate frames. The image resolution in this dataset is around $640\times 480$. Following previous methods, we report the average interpolation error~(IE) on the OTHER dataset. A lower IE indicates better performance.

	%\noindent\textbf{HD~\cite{bao2019memc}.} The videos in this dataset are of high-resolution and the motions are typically large. It contains four $1920\times 1080$p, three $1280\times 720$p and four $1280\times 544$p videos. Following the author of HD benchmark, we use the first 100 frames of each video for evaluation(TODO: words from RIFE).
	\vspace{0.05in}	
	\noindent\textbf{SNU-FILM~\cite{snufilm}.} It contains $1,240$ triplets of resolutions up to $1280\times 720$. There are four different settings according to the motion types: Easy, Medium, Hard and Extreme.

	\begin{table*}[t]
		\setlength{\belowcaptionskip}{0pt}
		\scalebox{0.95}{
			\begin{tabular}{l|ccccccc}
				\hline 
				\multicolumn{1}{c|}{\multirow{2}{*}{Method}} & \multicolumn{1}{c}{\multirow{2}{*}{Vimeo90K}} & \multicolumn{1}{c}{\multirow{2}{*}{UCF101}} & \multicolumn{1}{c}{\multirow{2}{*}{Middlebury}} & \multicolumn{4}{c}{SNU-FILM} \\
				\cline{5-8} & & & & Easy & Medium & Hard & Extreme \\

				\hline 
				ToFlow~\cite{vimeo90k} & 33.73/0.9682 & 34.58/0.9667 & 2.15 & 39.08/0.9890 & 34.39/0.9740 & 28.44/0.9180 & 23.39/0.8310 \\ 
				SepConv~\cite{niklaus2017video} & 33.79/0.9702 & 34.78/0.9669 & 2.27 & 39.41/0.9900 & 34.97/0.9762 & 29.36/0.9253 & 24.31/0.8448 \\ 
				CyclicGen~\cite{liu2019deep} & 32.09/0.9490 & 35.11/0.9684 & - & 37.72/0.9840 & 32.47/0.9554 & 26.95/0.8871 & 22.70/0.8083 \\ 
				DAIN~\cite{bao2019depth} & 34.71/0.9756 & 34.99/0.9683 & 2.04 & 39.73/0.9902 & 35.46/0.9780 & 30.17/0.9335 & 25.09/0.8584 \\ 
				CAIN~\cite{snufilm} & 34.65/0.9730 & 34.91/0.9690 & 2.28 & 39.89/0.9900 & 35.61/0.9776 & 29.90/0.9292 & 24.78/0.8507 \\ 
				AdaCoF~\cite{lee2020adacof} & 34.47/0.9730 & 34.90/0.9680 & 2.24 & 39.80/0.9900 & 35.05/0.9754 & 29.46/0.9244 & 24.31/0.8439 \\ 
				BMBC~\cite{park2020bmbc} & 35.01/0.9764 & 35.15/0.9689 & 2.04 & 39.90/0.9902 & 35.31/0.9774 & 29.33/0.9270 & 23.92/0.8432 \\ 
				RIFE-Large~\cite{rife} & 36.10/0.9801 & 35.29/0.9693 & {\color{blue} 1.94} & {\color{blue} 40.02/0.9906} & {\color{blue} 35.92/0.9791} & 30.49/0.9364 & 25.24/0.8621 \\ 
				ABME~\cite{asymmetric} & {\color{blue} 36.18/0.9805} & {\color{blue} 35.38/0.9698} & 2.01 & 39.59/0.9901 & 35.77/0.9789 & {\color{blue} 30.58/0.9364} & {\color{blue} 25.42/0.8639} \\ 
				\textbf{Ours} & {\color{red} 36.50/0.9816} & {\color{red} 35.43/0.9700} & {\color{red} 1.82} & {\color{red} 40.13/0.9907} & {\color{red} 36.09/0.9799} & {\color{red} 30.67/0.9378} & {\color{red} 25.43/0.8643}  \\
				\hline 
		\end{tabular}}
		\caption{Quantitative comparison among different VFI methods on 4 testing datasets. We report the average interpolation error IE (the lower the better) on the Middlebury dataset and report PSNR/SSIM~(the higher the better) on other datasets. The best and second-best results are colored in {\color{red} red} and {\color{blue} blue}.}
		\label{table:quantitative}
		%\vspace{-0.10in}
	\end{table*}

	\begin{table*}[t]
		\setlength{\belowcaptionskip}{0pt}
		\centering
		\scalebox{1}{
			\begin{tabular}{l|cc|ccccc}
				\hline

				\multicolumn{1}{c|}{\multirow{2}{*}{ }} & \multicolumn{1}{c}{\multirow{2}{*}{TFLs}} & \multicolumn{1}{c}{\multirow{2}{*}{CSWA}} & \multicolumn{1}{c}{\multirow{2}{*}{Vimeo90K}} & \multicolumn{4}{c}{SNU-FILM} \\
				\cline{5-8} & & & & Easy & Medium & Hard & Extreme \\

				\hline 
				Model 1 & \xmark & \xmark & 36.27/0.9809 & 40.01/0.9906 & 35.89/0.9793 & 30.58/0.9369 & 25.33/0.8629 \\ 
				Model 2 & \cmark & \xmark & 36.49/0.9815 & 40.06/0.9907 & 36.03/0.9798 & 30.61/0.9375 & 25.40/0.8643  \\ 
				Model 3 & \cmark & \cmark & 36.50/0.9816 & 40.13/0.9907 & 36.09/0.9799 & 30.67/0.9378 & 25.43/0.8643  \\ 
				\hline 
		\end{tabular}}
		\caption{Ablation study on the proposed modules. }
		\label{table:aba_modules}
		%\vspace{-0.10in}
	\end{table*}

	\subsection{Implementation Details}
	
	\noindent\textbf{Network Achitecture.} In the VFIformer, the window size is set to $8\times 8$, the channel numbers of linear layers and convolution layers are 180. Each TFB contains 6 TFLs except that the first TFB only contains 2 TFLs. 
	Encoder $Enc$ contains 4 blocks and each extracts one level of features from $I_0$ and $I_1$. Each encoder block consists of 2 convolutions with strides 2 and 1, respectively, and the channel numbers of features are 24, 48, 96, and 192 from shallow to deep layers. 
	The architecture of the flow estimator is included in the supplementary file.

		\vspace{0.05in}
	\noindent\textbf{Training Details.} We train our model with the AdamW optimizer. The learning rate is set to $1e-4$. We first train the flow estimator for 0.32M iterations with batch size 48. Then the whole model is trained in an end-to-end manner for 0.47M iterations with batch size 24.
	The weight coefficients ${\lambda}_{rec}$, ${\lambda}_{css}$, and ${\lambda}_{dis}$ are 1, 1 and 0.01, respectively. We randomly crop $192\times 192$ patches from the training samples and augment them by random flip and time reversal.

	%------------------------------------------------------------------
	
	\subsection{Comparisons with State-of-the-Art Methods}
	
	\begin{figure*}[t]
		\begin{center}
			\includegraphics[width=1.0\linewidth]{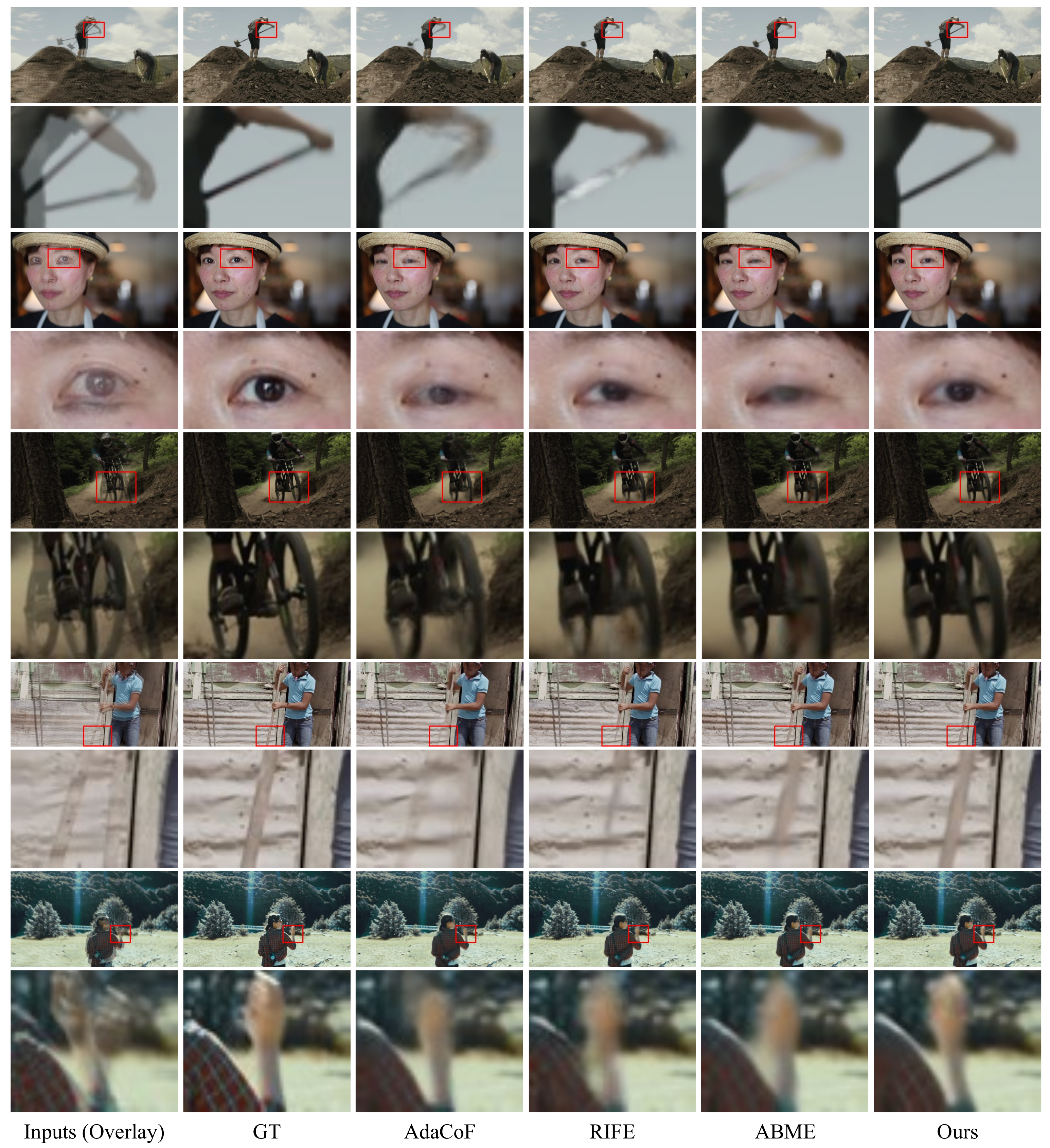}
		\end{center}
		\vspace{-0.15in}
		\caption{Visual comparison among different VFI methods on the Vimeo90K testing set.}
		\label{fig:qualitative}
		\vspace{-0.1in}
	\end{figure*}

	\begin{figure*}[t]
		\begin{center}
			\includegraphics[width=1.0\linewidth]{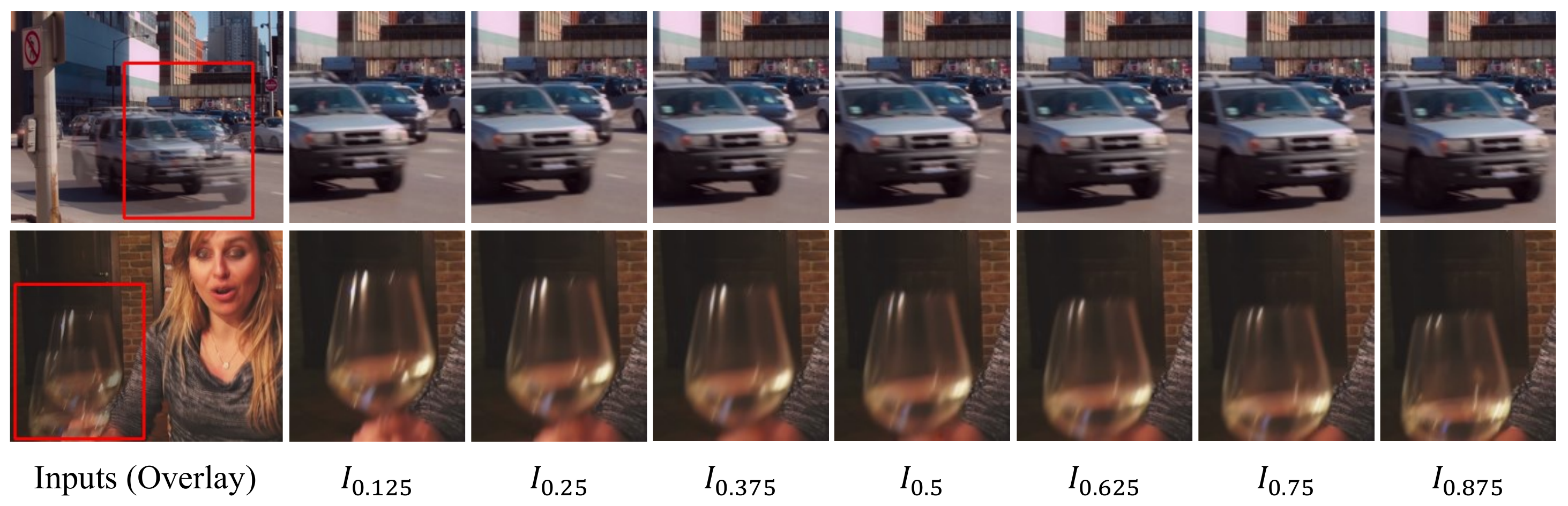}
		\end{center}
		\vspace{-0.1in}
		\caption{$8 \times$ interpolation results of our method on the Vimeo90K testing set.}
		\label{fig:multi_frame}
	\end{figure*}

	\begin{figure}[t]
		\begin{center}
			\includegraphics[width=1.0\linewidth]{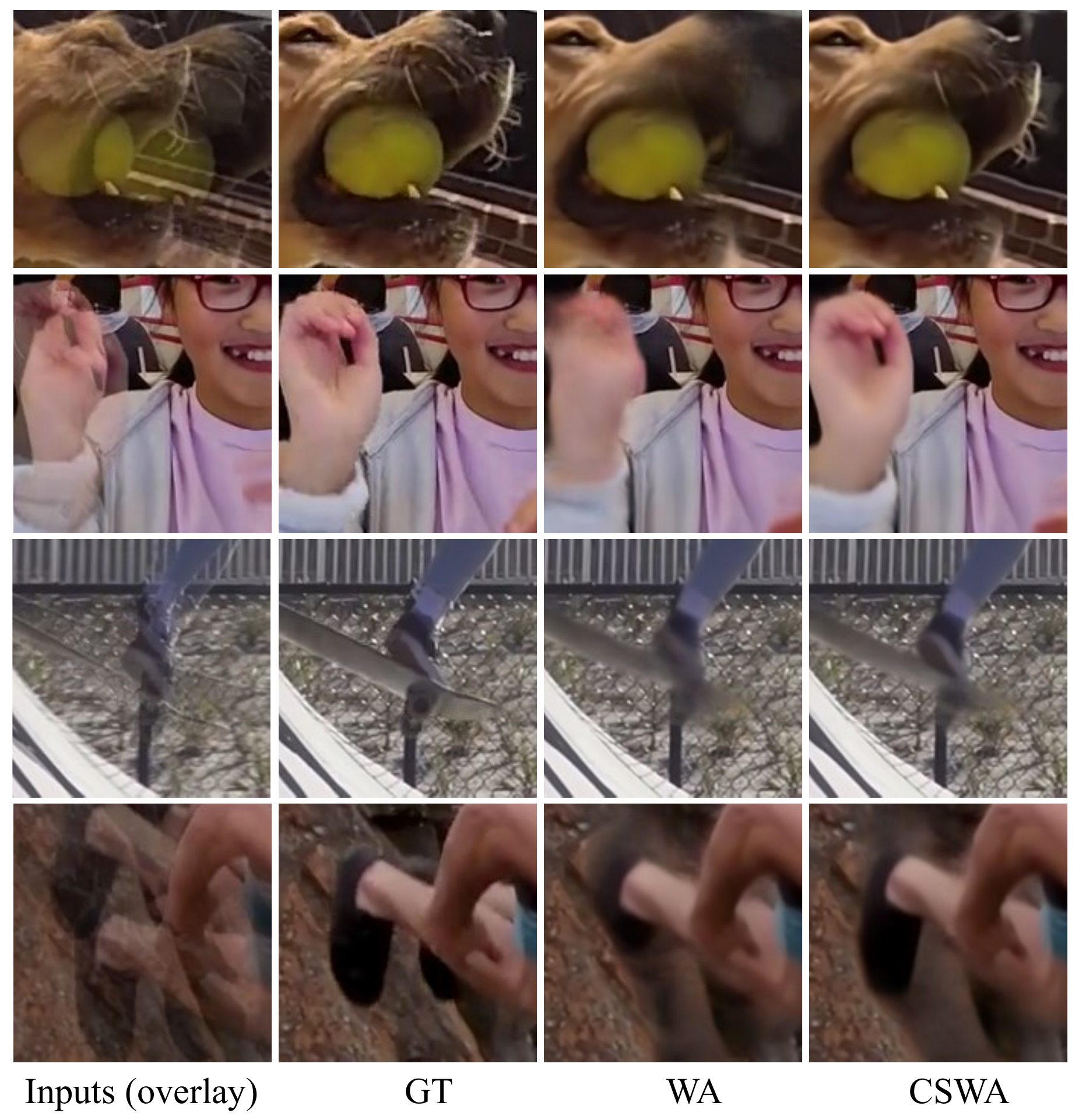}
		\end{center}
		\vspace{-0.05in}
		\caption{Ablation study on the cross-scale window-based attention~(CSWA), the samples in the first and second columns are overlaid inputs and ground-truths. Those in the third and fourth are the results of the models without and with CSWA, respectively.}
		\label{fig:aba_cswa}
		\vspace{-0.1in}
	\end{figure}
	
	We compare our model with nine recent, competitive methods, including ToFlow~\cite{vimeo90k}, SepConv~\cite{niklaus2017video}, CyclicGen~\cite{liu2019deep}, DAIN~\cite{bao2019depth}, CAIN~\cite{snufilm}, AdaCoF~\cite{lee2020adacof}, BMBC~\cite{park2020bmbc}, RIFE-Large~\cite{rife} and ABME~\cite{asymmetric}. Table~\ref{table:quantitative} shows the quantitative comparison, where the best and second best results are colored in red and blue. It is observed that our model outperforms recent state-of-the-art methods on all four testing sets.
	It is noteworthy that our method outperforms the second-best method on Vimeo90K testing set by \textbf{0.32} dB.
	
	The visual comparison between our method and other VFI methods is shown in Fig.~\ref{fig:qualitative}. Our proposed method generates more reasonable results with fewer unpleasing artifacts in general. For example, our method successfully interpolates the intermediate frame of the stick with large motion in the first and fourth example of Fig.~\ref{fig:qualitative}.
	
	Meanwhile, to thoroughly investigate the performance of our proposed method, we also conduct multi-frame generation. We recursively apply our model to generate multiple intermediate frames. Specifically, given two input frames $\bm{I}_0$ and $\bm{I}_1$, we first generate $\bm{I}_{0.5}$. Then we interpolate between $\bm{I}_0$ and $\bm{I}_{0.5}$ to generate $\bm{I}_{0.25}$. We show the $8 \times$ interpolation results on Vimeo90K testing set in Fig.~\ref{fig:multi_frame}. Our model yields multiple intermediate frames with smooth motion.

	%------------------------------------------------------------------------

	\begin{table}
		\setlength{\belowcaptionskip}{0pt}
		\centering
		\scalebox{1}{
			\begin{tabular}{c|c}
				\hline 
				Window Size & ~~~~~~~PSNR/SSIM~~~~~~~ \\ 
				\hline
				4 & ~~~~~~~36.24/0.9806~~~~~~~  \\
				8 & ~~~~~~~36.29/0.9807~~~~~~~  \\
				12 & ~~~~~~~36.31/0.9808~~~~~~~  \\
				\hline 
		\end{tabular}}
		\caption{Ablation study on the window sizes of self-attention. }
		\label{table:aba_ws}
		\vspace{-0.10in}
	\end{table}

	\subsection{Ablation Study}

	In this section, we conduct several ablation studies to investigate our proposed method. We verify the effectiveness of the Transformer layers and the proposed cross-scale window-based attention. We also analyze the influence of different window sizes while computing attention.
	
		\vspace{0.05in}
	\noindent\textbf{Effect of the Transformer layers~(TFLs).} TFLs are the key components of our VFIformer, which play the role of capturing long-range dependency. We investigate the effect of TFLs by replacing them with convolutional layers of a similar number of parameters. The ablation results are shown in Table~\ref{table:aba_modules}, where Model 2 is the model with TFLs~(using standard window-based attention) and Model 1 is the model without TFLs.
	Model 2 outperforms Model 1 by 0.22 dB on the Vimeo90K testing set, and also obtains better performance on the SNU-FILM dataset under 4 settings.

		\vspace{0.05in}
	\noindent\textbf{Effect of Cross-scale Window-based Attention. } Cross-scale window-based attention~(CSWA) is proposed to enlarge the receptive field and aggregate multi-scale information. 
	To further verify its effectiveness, we train two models with and without CSWA respectively. As shown in Table~\ref{table:aba_modules}, Model 2 adopts standard window-based attention~(WA) while Model 3 adopts CSWA.
	
	Compared with Model 2, Model 3 improves it by 0.07, 0.06, and 0.06 dB in the Easy, Medium, and Hard settings of SNU-FILM, respectively, in terms of PSNR. 
	We show the visual comparison of these two models on SNU-FILM~(Hard) in Fig.~\ref{fig:aba_cswa}. It is observed that compared with Model 2, Model 3 generates sharper results with more fine details while dealing with cases with large motion.
	%To have a thorough investigation on the Vimeo90K testing set, we split it into 4 motion levels according to the average flow values estimated by PWC-Net~\cite{pwcnet}. The PSNR gain of CSWA becomes larger as motions become larger as shown in Table~\ref{table:motion}, which further validates the effectiveness of CSWA.

		\vspace{0.05in}
	\noindent\textbf{Influence of the Attention Window Size. } We further investigate the influence of attention window size, which determines the size of the receptive field. The larger the window is, the larger the range of information can be captured, along with higher computational cost. To enable quick exploration, we train all the models for only 0.36M iterations in this experiment. Table~\ref{table:aba_ws} shows the ablation result. It is observed that the model with window size 12 achieves the best performance. To balance the performance and the computational cost, we choose window size 8 in our experiments.

	%------------------------------------------------------------------------
	\section{Limitations}
	
	Though our proposed method has achieved decent results, there are several limitations. First, while our model is built upon the window-based attention, the computational cost is still heavier than CNN-based methods due to the complex calculations of self-attention. We will explore more efficient approaches in the future by computing self-attention in the horizontal and vertical stripes in parallel~\cite{dong2021cswin}. 
	
	Second, unlike existing methods~\cite{jiang2018super,asymmetric} that are able to interpolate frames at arbitrary time, our model only synthesizes the intermediate frame. In future work, we will investigate variables that represent the interpolation time and put them into the network to control the generated content.

	\section{Conclusion}
	
	In this work, we have proposed a novel framework integrated with the Transformer for the video frame interpolation task. The proposed VFIformer endows our framework with a strong capability of modeling long-range dependencies and handling large motions. Further, a novel cross-scale window-based attention mechanism is designed to aggregate multi-scale information and enlarge the receptive field. Extensive experiments show that our proposed method achieves superior performance over existing state-of-the-art methods on multiple popular benchmarks.

	%%%%%%%%% REFERENCES
	{\small
		\bibliographystyle{ieee_fullname}
		\bibliography{egbib}
	}

	%------------------------------------------------------------------------
	\clearpage
	\newpage
	
	\appendix
	
	\section{Appendix}
	
	\subsection{More Implementation Details}
	\label{sec:imp}

	\begin{figure}[h]
		\begin{center}
			\includegraphics[width=1.0\linewidth]{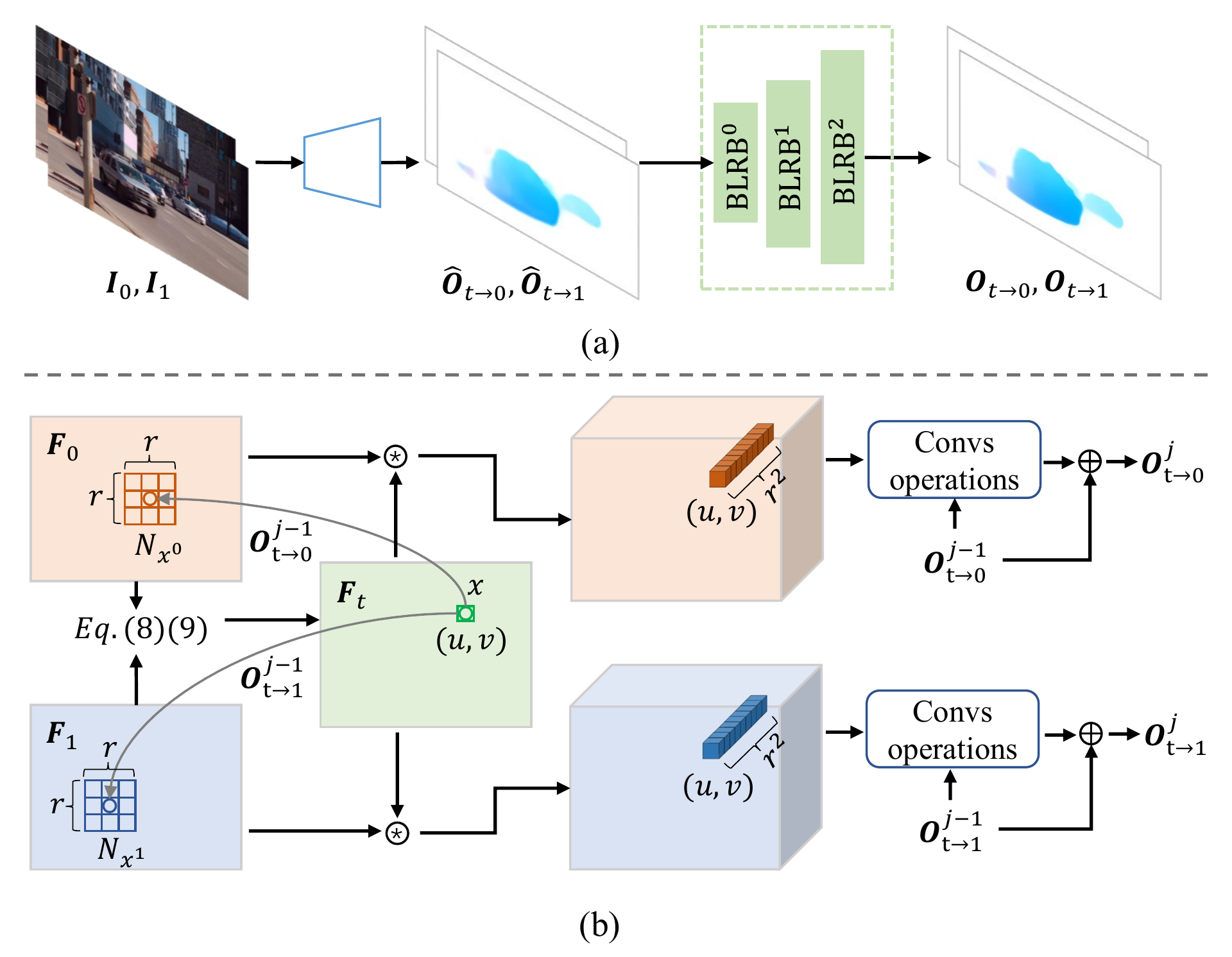}
		\end{center}
		\vspace{-0.1in}
		\caption{(a). The pipeline of our flow estimator, where the Bilateral Local Refinement Block~(BLRB) is used to refine the flows. (b). Details of the Bilateral Local Refinement Block.}
		\label{fig:flow_estimator}
		%\vspace{-0.1in}
	\end{figure}

	\noindent\textbf{Flow Estimator Achitecture.} The pipeline of our flow estimator is shown in Fig.~\ref{fig:flow_estimator}a, a flow prediction network~\cite{rife} is first used to predict the coarse flows $\hat{\bm{O}}_{t\rightarrow0}$, $\hat{\bm{O}}_{t\rightarrow1}$. Then Bilateral Local Refinement Blocks~(BLRBs) are used to refine the flows in a coarse-to-fine manner, whose details are shown in Fig.~\ref{fig:flow_estimator}b.
	
	For the $j$-th BLRB, given the flows $\bm{O}^{j-1}_{t\rightarrow0}$ and $\bm{O}^{j-1}_{t\rightarrow1}$ produced by the last BLRB and feature maps $\bm{F}_0^{j}$, $\bm{F}_1^{j}$ of two input frames extracted by the encoder $Enc$, we first rescale the flows to the current scale~(note that we do not change their notations after rescaling for brevity) and use them to backward warp the features, obtaining $\widetilde{\bm{F}}_0^{j}$ and $\widetilde{\bm{F}}_1^{j}$. Then the warped features are fed into convolutional layers to produce the intermediate feature $\bm{F}_t^{j}$. The specific process is given by
	\begin{align}\label{eq:flow}
	\bm{D}=sigmoid(Convs([\widetilde{\bm{F}}_0^{j}, \widetilde{\bm{F}}_1^{j}])) \;, \\
	\bm{F}_t^{j} = \bm{D} \odot \widetilde{\bm{F}}_0^{j}+(1-\bm{D}) \odot \widetilde{\bm{F}}_1^{j} \;,
	\end{align}
	where $Convs$ denotes convolutional layers, $[~]$ denotes concatenation in the channel dimension, $\bm{D}$ is the generated mask for blending $\widetilde{\bm{F}}_0^{j}$ and $\widetilde{\bm{F}}_1^{j}$, and $sigmoid$ is used to ensure the mask in the range of [0, 1].
	
	Afterwards, we compute the local correlation volumes to model the relationships among pixels in $\bm{F}_t^{j}$ and $\bm{F}_0^{j}$, $F_1^{j}$. Specifically, for each pixel $x=(u,v)$ in $\bm{F}_t^{j}$, we map it to its estimated correspondence in $\bm{F}_0^{j}$~(here we take refining $\hat{\bm{O}}^{j}_{t\rightarrow0}$ for example): $x^{'}=(u+\hat{\bm{O}}^{j-1}_{t\rightarrow0}(u), v+\hat{\bm{O}}^{j-1}_{t\rightarrow0}(v))$. Then a local window around $x^{'}$ is defined as
	\begin{align}
	\mathcal{N}_{x^{'}} = \left \{ (u+d_u, v+dv)~|~d_u,d_v\in \left \{ -r,...,r \right \} \right \} \;,
	\end{align} 
	where $r$ is the radius and is set to $1$ in our experiments. Then we calculate the cosine similarity between $x$ and pixels in $\mathcal{N}_{x^{'}}$, producing a correlation vector.
	The correlation map is then processed by two convolutional layers. Meanwhile, the flow $\bm{O}^{j-1}_{t\rightarrow0}$ is also processed with other two convolutional layers. At last, the correlation feature, the flow feature, $\bm{F}_0$ and $\bm{F}_t$ are concatenated and fed into four convolutional layers to produce the flow residual $\Delta \bm{O}^{j}_{t\rightarrow0}$, the refined flow is obtained as
	\begin{align}
	\bm{O}^{j}_{t\rightarrow0} = \Delta \bm{O}^{j}_{t\rightarrow0} + \bm{O}^{j-1}_{t\rightarrow0} \;.
	\end{align} 
	Note that all the operations are similar for $\bm{O}^{j}_{t\rightarrow1}$.

	%--------------------------------------------------------------------

	\subsection{More Ablation Studies}
	
	\begin{table}[h]
		\setlength{\belowcaptionskip}{0pt}
		\centering
		\scalebox{1}{
			\begin{tabular}{c|c}
				\hline 
				TFL number & ~~~~~~~PSNR/SSIM~~~~~~~ \\ 
				\hline
				2 & ~~~~~~~36.35/0.9810~~~~~~~  \\
				4 & ~~~~~~~36.43/0.9814~~~~~~~  \\
				6 & ~~~~~~~36.49/0.9815~~~~~~~  \\
				8 & ~~~~~~~36.52/0.9817~~~~~~~  \\
				\hline 
		\end{tabular}}
		\caption{Ablation study on the TFL number. }
		\label{table:aba_tflnb}
		%\vspace{-0.10in}
	\end{table}

	\noindent\textbf{Influence of the TFL number.} We also investigate the influence of the TFL number, the ablation results on the Vimeo90K~\cite{vimeo90k} testing set are shown in Table~\ref{table:aba_tflnb}, we set the numbers of the TFLs in each TFB to 2, 4, 6, and 8, respectively. It is observed that the more TFLs, the higher PSNR/SSIM. To balance the performance and the computational cost, we set the number to 6 in our experiments.
	
	\noindent\textbf{Effect of CSWA.} To have a thorough investigation on the Vimeo90K testing set, we split it into 4 motion levels according to the average flow values estimated by PWC-Net~\cite{pwcnet}. The PSNR gain of CSWA becomes larger as motions become larger as shown in Table~\ref{table:motion}, which further validates the effectiveness of CSWA. 
	
	\begin{table}[h]
		\setlength{\belowcaptionskip}{0pt}
		\centering
		\scalebox{1}{
			\begin{tabular}{|cc|cc|c|}
				\hline
				flow range & avg. flow &  WA & CSWA & gain \\
				\hline
				%$1.9$ & 36.51 & 36.52 & 0.01  \\
				%$4.0$ & 36.52 & 36.55 & 0.03 \\
				%$7.4$ & 36.91 & 36.94 &  0.03 \\
				%$14.02$ & 34.00 & 34.04 & 0.04 \\
				$[0,3)$ & $1.9$&  36.51 & 36.52 & 0.01  \\
				$[3,6)$ & $4.0$ & 36.52 & 36.55 & 0.03 \\
				$[6,10)$ & $7.4$ & 36.91 & 36.94 &  0.03 \\
				$[10,+\infty)$ & $14.02$ & 34.00 & 34.04 & 0.04 \\
				\hline
		\end{tabular}}
		\captionof{table}{PSNR of WA and CSWA under 4 motion levels.}
		\label{table:motion}
		%\vspace{-0.10in}
	\end{table}
	
	\subsection{More Quantitative Comparisons}
	
	\begin{table}[h]
		\setlength{\belowcaptionskip}{0pt}
		\centering
		\footnotesize
		\scalebox{0.9}{
			\begin{tabular}{l|cccc}
				\hline 
				Methods & Runtime~(ms) & Param.~(M) & Vimeo90K \\
				\hline 
				SoftSplat-$\mathcal{L}$~\cite{softsplat} & - & -  & 36.10/0.9700 \\
				CDFI~\cite{cdfi} & 60 & 5.0 & 35.17/0.9640 \\
				XVFI~\cite{xvfi} & 96 & 5.5 & 35.07/0.9760 \\
				BMBC~\cite{park2020bmbc} & 478 & 11.0 & 35.01/0.9764 \\
				RIFE-Large~\cite{rife} & 13 & 21.7 & 36.10/0.9801 \\
				ABME~\cite{asymmetric} & 158 & 18.1 & 36.18/0.9805 \\
				Ours & 365 & 24.1 & 36.50/0.9816 \\
				Ours-Small & 213 & 17.0 & 36.38/0.9811 \\
				\hline 
		\end{tabular}}
		%\vspace{-0.14in}
		\caption{More quantitative comparisons. The running time is tested on images with $448\times 256$ resolution on an NVIDIA TITAN V GPU.}
		\label{table:runtime_comparison}
		%\vspace{-0.16in}
	\end{table}
	
	We compare the running time and network parameters with some recent SOTA methods in Table~\ref{table:runtime_comparison}.
	As mentioned in the \textit{Limitation} section of the main paper, the computational cost of our model is still heavier than CNN-based methods. To increase the practicability of our model, we further train a light-weight version, which is denoted as `Ours-Small'. It adopts a simpler flow estimator architecture~\cite{rife} and its window size and channel number are $4\times 4$ and 136. As shown in Table~\ref{table:runtime_comparison}, the light-weight version has moderate running time and parameters yet still yields the SOTA result.

	%--------------------------------------------------------------------
	
	\subsection{More Visual Results}
	
\begin{figure*}[t]
		\begin{center}
			\includegraphics[width=1.0\linewidth]{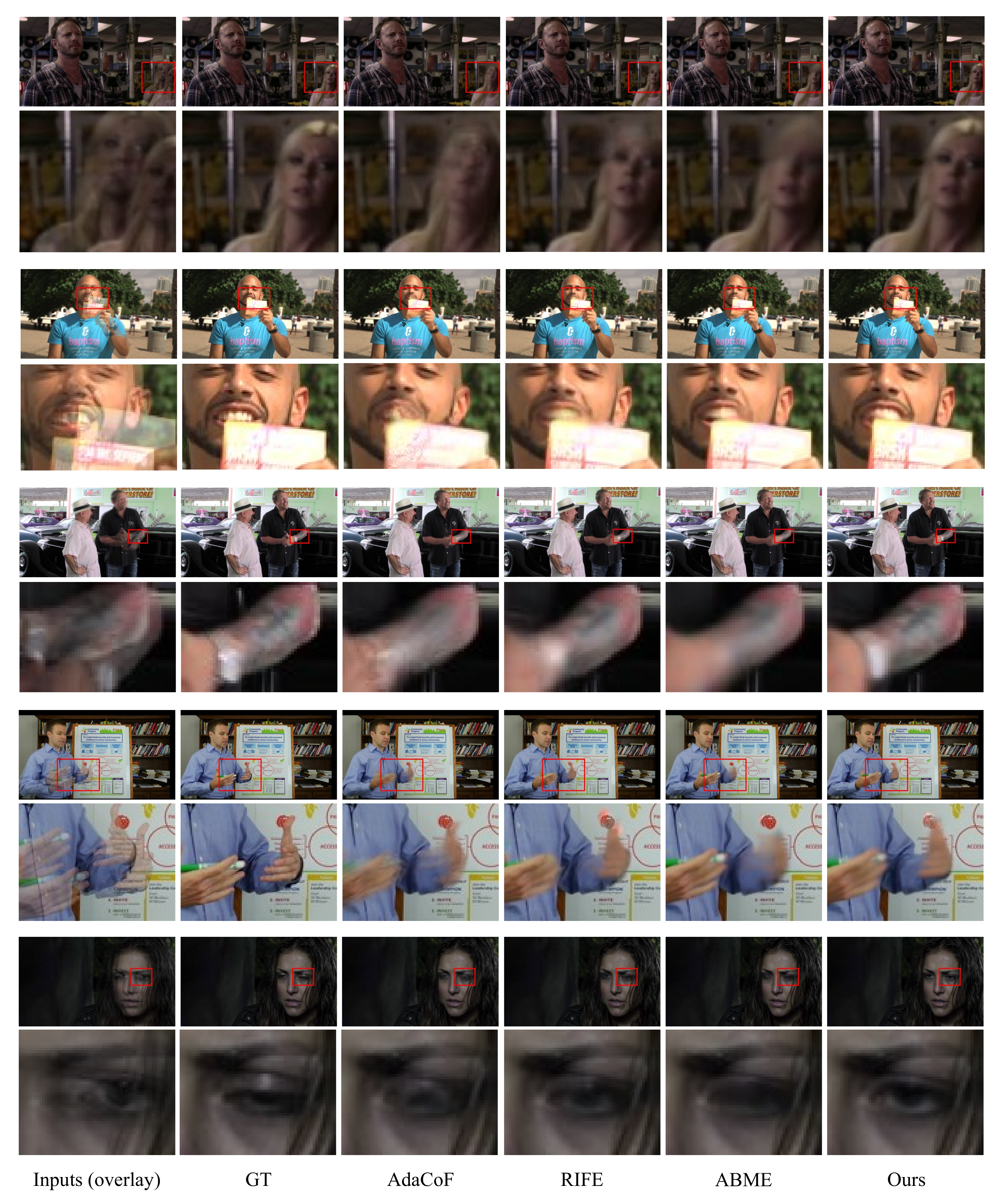}
		\end{center}
		\vspace{-0.15in}
		\caption{Visual comparison among different VFI methods on the Vimeo90K~\cite{vimeo90k} testing set.}
		\label{fig:comparison_supp_0}
		%\vspace{-0.1in}
	\end{figure*}

\begin{figure*}[t]
		\begin{center}
			\includegraphics[width=1.0\linewidth]{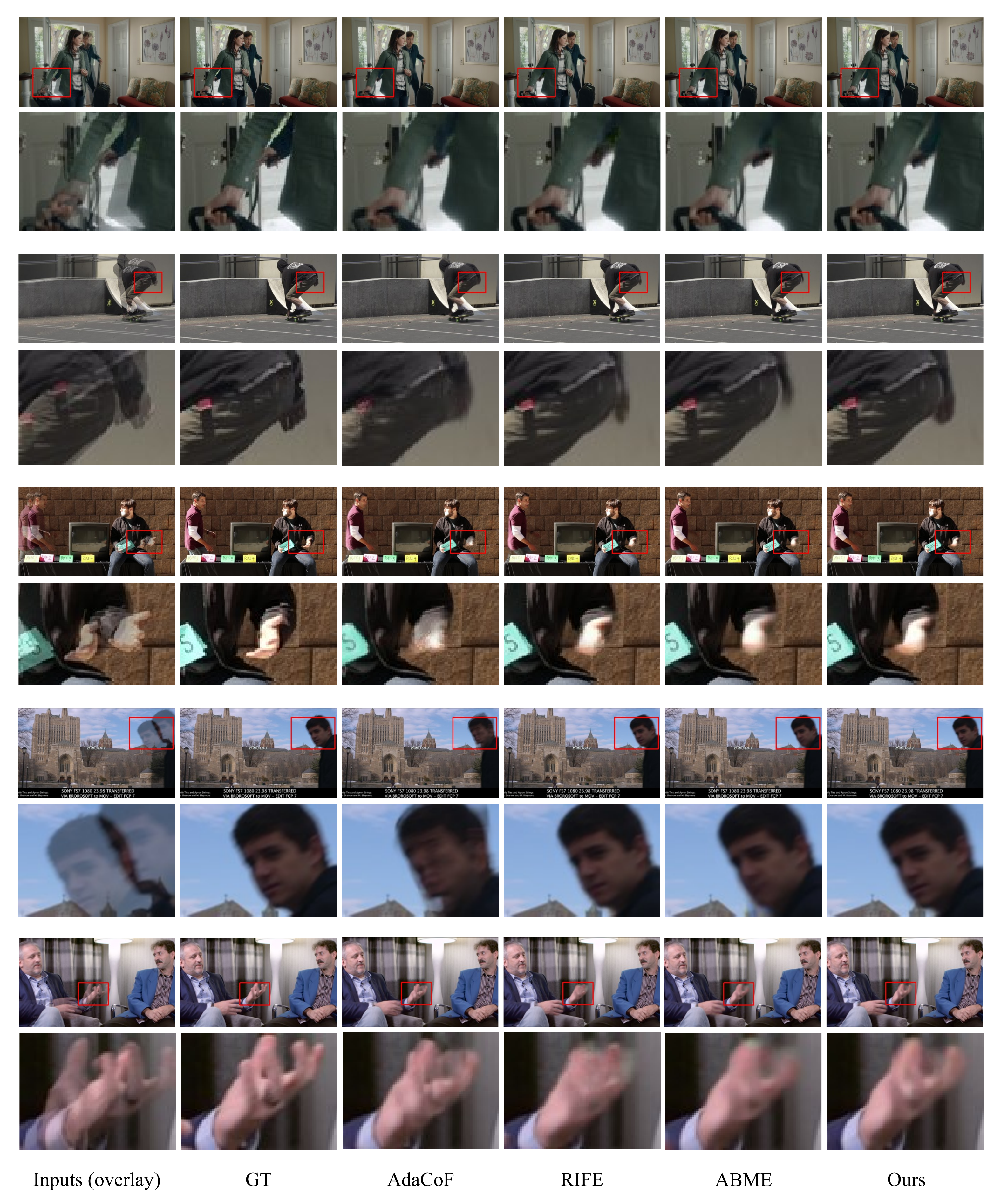}
		\end{center}
		\vspace{-0.15in}
		\caption{Visual comparison among different VFI methods on the Vimeo90K~\cite{vimeo90k} testing set.}
		\label{fig:comparison_supp_1}
		%\vspace{-0.1in}
	\end{figure*}

	More visual results are shown in Fig.~\ref{fig:comparison_supp_0} and Fig.~\ref{fig:comparison_supp_1}.
	We compare our proposed method and other recent state-of-the-art methods, including AdaCoF~\cite{lee2020adacof}, RIFE-Large~\cite{rife} and ABME~\cite{asymmetric}. It can be observed that our method restores more appealing results with sharper structures. 
	%We also include some video examples in \textit{examples.mp4}.

\end{document}